# Stingy Context: 18:1 Hierarchical Code Compression for LLM Auto-Coding

v5

David Linus Ostby
Chief Science Officer, ViperPrompt
stingycontext@viperprompt.ai

December 2025

## Abstract

We introduce **Stingy Context**, a hierarchical tree-based compression scheme achieving 18:1 reduction in LLM context tokens for auto-coding tasks. Using our **TREEFRAG** exploit decomposition, we reduce a real source code base of 239k tokens to 11k tokens while preserving task fidelity. Empirical results across 12 Frontier models show 94 to 97% success on 40 real-world issues at low cost, outperforming flat methods and mitigating lost-in-the-middle effects.

---

## 1 Introduction

Large language models have scaled context windows to millions of tokens, yet real-world auto-coding tasks remain constrained by token limits, inference costs, and the "lost in the middle" phenomenon.

While labs tout ever-larger contexts, practical codebases exceed even 200k windows when serialized naively, forcing developers into brittle chunking or incomplete prompts.

We present **Stingy Context**: a hierarchical tree-based compression scheme achieving a conservative 18:1 token reduction for LLM-driven software evolution. By decomposing applications into a **TREEFRAG** tree structure (a homogenized hierarchical representation) we compress a real 20,756 line, 1,014-node code base (the **RWT**) from ~200k raw tokens to ~11k tokens while preserving full structural fidelity.

Empirical evaluation across 40 real-world tasks (bugs, enhancements, refactors) shows top models achieve 94 to 97% success at cents-per-task cost, outperforming three common auto-coding methods and enabling sustainable, iterative auto-coding on consumer hardware.

This work establishes hierarchical compression as a scalable alternative to brute-force context stuffing.

**Stingy Context** proves hierarchical compression is a smart alternative to ever-larger contexts.

## 2 Background

LLM context windows have ballooned to 1M+ tokens, but auto-coding suffers: 200k limits choke on 20k line apps, while "lost in the middle" halves accuracy mid-prompt. Flat chunking/RAG ignores hierarchy, yielding incoherent patches.

Prior fixes fail: **LongRoPE** extends context windows but is not designed to capture application architecture… LongRoPE embeddings lose structure. We need hierarchy-aware compression.

A key issue is the ”lost in the middle” phenomenon. Liu et al. (2024) show LLMs excel when relevant info is at prompt start/end but degrade up to 40% when buried in the middle, persisting in long-context models. In auto-coding, full code bases bury details, raising errors and ”creative” unwanted changes.

Currently, this U-shaped lost-in-the-middle performance curve persists across tasks like multi-document QA and key-value retrieval, affecting both open and closed models. In our use case of interest (auto-coding), naive full-codebase prompts dilute focus, raising costs and errors.

Flat approaches (chunking, RAG) ignore hierarchy, failing on dependencies. **Stingy Context** counters this with tree-based compression, keeping critical info at prompt peripheries.

Large language models now handle million-token contexts, but auto-coding - spec-driven generation where users describe needs and LLMs produce code - remains challenging. Traditional ”vibe coding” relies on developers writing code directly, but no-code systems delegate all coding to LLMs across language(s). This demands *just the right context*: enough for accuracy, minimal to cut costs and focus attention. Optimization of the context window is key.

Source code is linear: a one-dimensional string prone to overwhelming LLMs. Trees impose hierarchy, enabling navigation and higher-dimensional structure for compression (**TREEFRAG** exploits this for its >18:1 compression ratios).

Software development involves more than coding (30% effort); research, stakeholder engagement, and project management dominate. **TREEFRAG** aids in multiple areas of the software development process with its tree-based specs.

Flat methods (chunking, RAG) ignore hierarchy and dependencies. **TREEFRAG** counters by limiting exposure to function-level code, reducing overreach.

Trees as holders of structural knowledge is not a new concept. Tree structures abound in computing and engineering: file systems, DOM in HTML, database indexes (B-trees), GUI widgets, parse trees, decision trees. Later we will see how LLM tree-handling varies—empirical tables reveal capabilities.

Tree structures provide a human-readable and transparent method of understanding, documenting and accessing relevant components of an application quickly. **Stingy Context** leverages trees for focused, efficient LLM auto-coding.

## 3 Trees in the Real World

Hierarchical tree structures are ubiquitous in computing and everyday organization, enabling efficient navigation and representation of complex relationships in inherently flat data.

File systems organize directories and files recursively. Databases use B-trees for indexing, balancing search efficiency. GUI frameworks build widget trees: parents (windows, panels) contain children (buttons, labels). Parse trees represent code syntax. Decision trees model choices in ML.

Beyond computing: legal documents (sections/subsections), organizational charts, and project task breakdowns. In manufacturing: recursive bill of material, and exploded component views in blueprints, to name a few.

**Stingy Context** homogenizes these via a category-type table, unifying disparate hierarchies into one traversable tree for LLM reasoning and compression.

| Category | Type | Description |
|---|---|---|
| Project | Application | Root project node |
| Code | Python | Python source code node |
| GUI | Button | Clickable button |
| SQLite | Table | Database table |
| Specification | Vision | High-level vision node |
| Feature | Survey | User feedback survey node |

Table 1: Above is a short extract from a typical category-type table for homogenization.

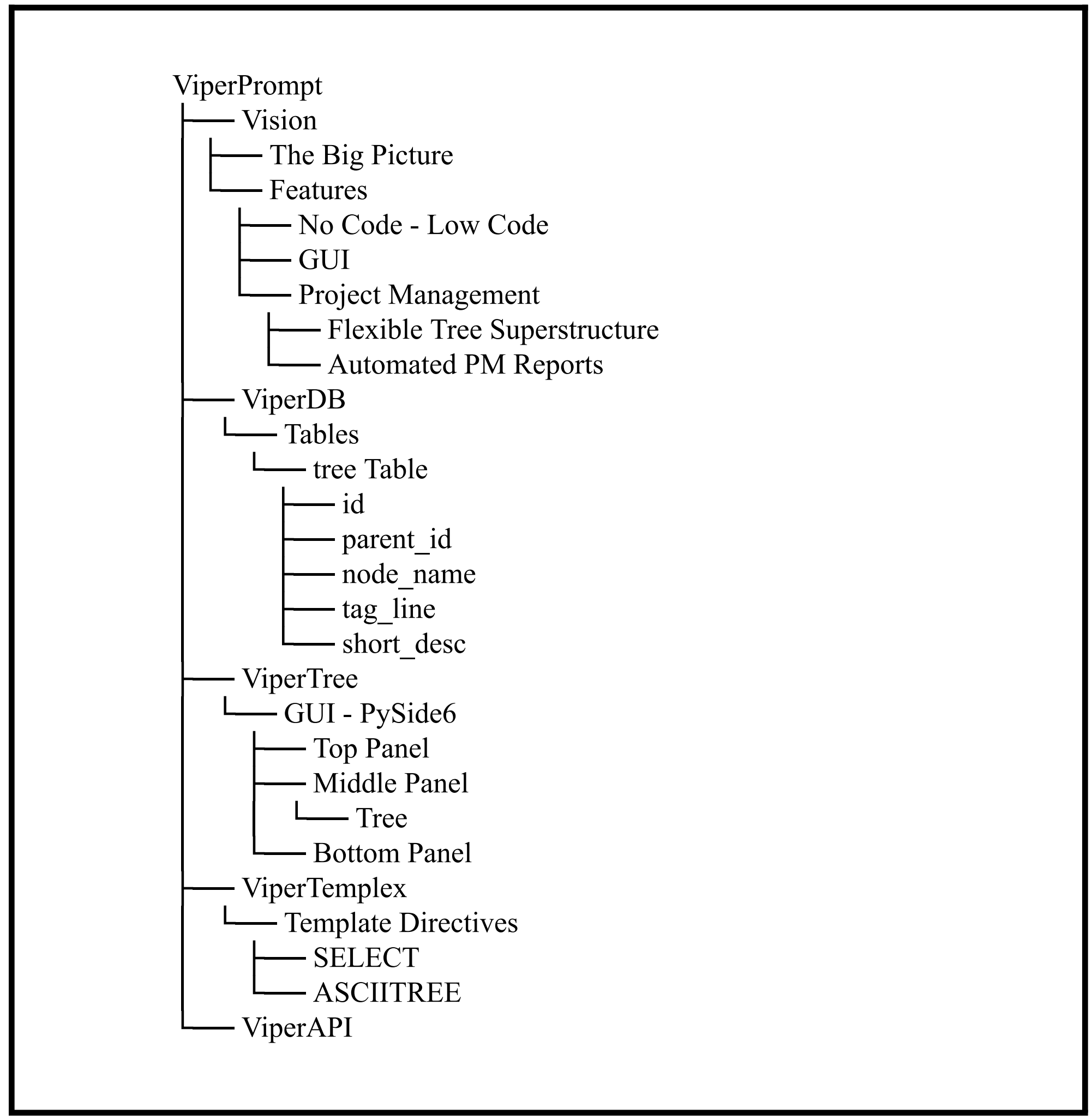

Figure 1: Simplified ASCII tree of the ViperPrompt RWT at low resolution.

## 4 LLM Model Token Counts for the RWT Raw Source Code Base

In this paper we will be using **RWT** (Real World Tree), a real, homogenized, 20,756 line pre-production application tree, See Appendix A for a detailed description of the **RWT** application and code base.

Below is a table giving counts for the tokenized **RWT** using our twelve MOI (Models Of Interest). We chose 12 MOI by taking three of the most current models across four LLM labs: Anthropic, Google, OpenAI and xAI.

| LLM Model of Interest | Tokens | Error |
|---|---|---|
| **claude-opus-4-5** | 210,519 | 400 error: prompt is too long : 210519 tokens > 200000 max |
| **claude-sonnet-4-5** | 210,519 | 400 error: prompt is too long : 210519 tokens > 200000 max |
| **claude-haiku-4-5** | 210,514 | 400 error: prompt is too long : 210514 tokens > 200000 max |
| **gemini-3-pro-preview** | 239,154 | |
| **gemini-2.5-pro** | 239,154 | |
| **gemini-2.5-flash** | 239,154 | |
| **gpt-5.2** | 196,361 | |
| **gpt-5.1** | 196,361 | |
| **gpt-4.1** | 196,361 | |
| **grok-4-1-fast-reasoning** | 195,451 | |
| **grok-4-1-fast-non-reasoning** | 195,451 | |
| **grok-code-fast-1** | 195,451 | |

Table 2: Token Counts for the **RWT** raw code base using different LLM models

These counts are important as they form the baseline numbers for computing compression ratios later in this paper. Note that although these models are tokenizing the same source code base (the **RWT** code base), their tokenization algorithms vary, producing varying token counts and, ultimately, compression ratios. Also note: the tokenized **RWT** codebase overruns the Anthropic model's context window size limits. Although this causes some trouble later in our experiments, we work around the issue so that we can obtain performance for Anthropic models in spite of this limitation..

Throughout this paper, we will be displaying various model token counts. These token counts evaluate to various compression ratios, affecting AI agent algorithmic feasibility and inference costs.

## 5 TREEFRAG: Hierarchical Decomposition and Compression

**TREEFRAG** is at the heart of our **Stingy Context** exploit. **TREEFRAG** decomposes codebases into function-level fragments stored as nodes in a hierarchical tree, mirroring software architecture. The decomposition process is not the subject of this paper. However it is a fairly mechanical process, especially with LLM assistance. In the case of our notional code base (the **RWT** application), the **TREEFRAG** decomposition is an automated and on-going process, utilizing LLMs for all the heavy decomposition activities.

Each type of element (e.g. a GUI) in the software system needs to be decomposed into a subtree, and all the disparate subtrees can then be blended into one unified **TREEFRAG** tree using the category-type table as the homogenizing backbone. By way of illustration, the complete category - type table for a typical application (such as **RWT**) is given in Appendix B. It is user extensible, providing for a wide variety of sub-tree homogenization use cases: here we focus on the auto-coding use case.

Trees can convey structure at all scales… from 'big picture' features down to the smallest level of detail. We will see later that, given a **TREEFRAG** tree object, LLMs have an uncanny ability to understand software systems with high accuracy. This includes the ability to locate not only the specific location of areas-of-interest, but also surmise likely solutions to architecturally-driven problems. Whether this is an 'emergent ability' is not of concern to our thesis, but nevertheless it is an unexpected skill, at least to the level we document below.

In **TREEFRAG**, we choose our levels of detail to match typical software architectures. Each node in a **TREEFRAG** tree holds metadata (name, category, type, tag line, short description, commentary, code) unified via the category-type table. This metadata is generated by parsing / summarizing application code using LLM single shots, and each metadata field is filled with decomposition material: tree nodes document their application counterparts in increasing LOD (Levels Of Detail). In this way, a **TREEFRAG** tree is a surrogate for the application.

Except at the highest LOD (7), raw code bodies are excluded from a **TREEFRAG** tree dump; only metadata is serialized as an ASCII tree, CSV, JSON, etc.

This paper focuses on LOD level 1, but we provide the table below to document the LOD concept for follow-on explorations:

| LOD | Level of Detail Description |
|---|---|
| **7** | Full raw source code from the application code base. |
| **6** | Commentary - a multi-paragraph summary of the node. |
| **5** | Short description - a one or two sentence description of the node. |
| **4** | Tag line - a title, a phrase or a few words describing the node. |
| **3** | Type - the 'type' code from the category / type table. |
| **2** | Category - the 'category' code from the category/type table. |
| **1** | Node name - a word or two, or a common abbreviation. |

Table 3: **TREEFRAG** LOD: seven levels of detail. These levels are arbitrary but importantly they are escalating in magnitude and detail with each increase in level.

The **TREEFRAG** process turns flat, one dimensional source code into a tree object which has a fractal dimension of 2. This 1D-to-2D shift yields 18:1 compression (at a minimum) for typical auto-coding use cases. So for our example **RWT** code base (20,756 lines, ~200k raw tokens), **TREEFRAG** yields a compression of the code base by a factor of 18.

These compressed fractal-like structures grow information density logarithmically with depth, and fights lost-in-the-middle effects not only by placing key context node meta data at prompt edges, but also by reducing the size of the 'middle' by over an order of magnitude. We do not know exactly why the **TREEFRAG** exploit is so effective, especially considering that a 18:1 compression ratio squeezes tokenized source code by 94%, but we surmise it is due to the shift from a 1D representation to a 2D tree object structure.

## 6 TREEFRAG Conveyance Formats

There are various formats useful to convey **TREEFRAG** trees to LLMs for processing. Different serialization formats yield varying compression ratios, as shown below. The 'gemini-3-pro-preview' model was used for this table. In practice, compression ratios will vary somewhat between models.

| Dump Format | TREEFRAG (tokens) | Code Base (tokens) | Comp Ratio | Description |
|---|---|---|---|---|
| **ASCII** | 11,358 | 239,153 | 21:1 | Human and LLM readable ASCII character visualization |
| **CSV** | 21,182 | 239,153 | 11:1 | Comma Separated Value Dump with a key list header |
| **JSON** | 39,390 | 239,153 | 6:1 | Typical JSON key value pairs |

Table 4: Conveyance Formats: LOD 1 **TREEFRAG** compression by format. Both raw code base tokens and **TREEFRAG** token counts vary depending on the LLM model. ASCII format has the highest compression ratio and also has the added benefit of being human readable.

## 7 Experiment 1: LLM Tree Capabilities

**TREEFRAG** sounds intriguing, but do LLMs understand tree structures well enough to be useful for the auto-coding use case? How good are LLMs at understanding tree structures?

To answer these questions we submit abstract 'theoretical' trees of various sizes and complexity to all of our models of interest, and then see whether these same models can answer useful questions about these trees.

The tree sizes are 50, 100, 250, 500, 750 and 1000 node trees. These trees are randomly created, and are abstract in the sense that the node names are simply a string of alphanumeric characters. As an illustration, here is a short excerpt from one of these trees shown below.

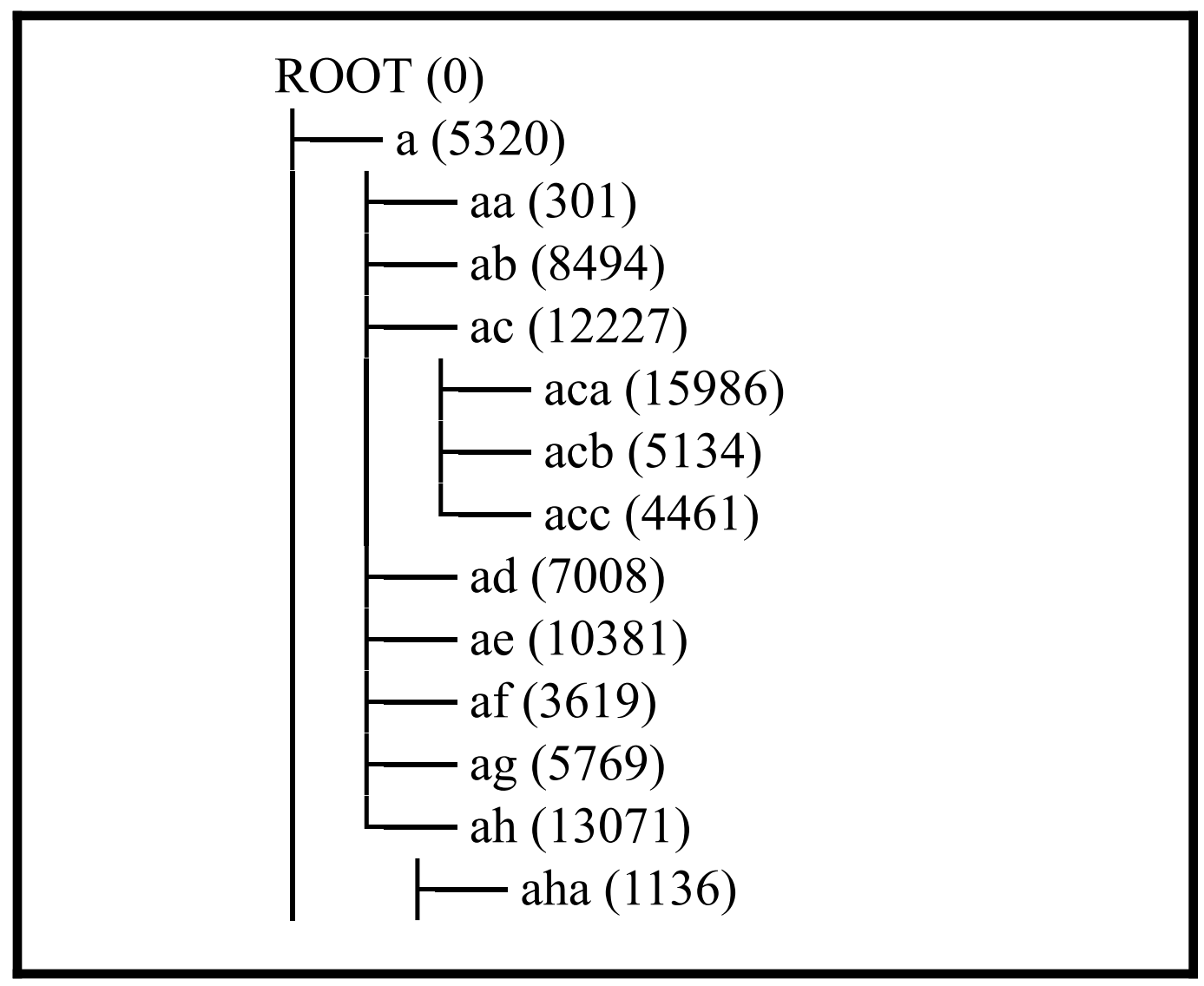

Figure 2: The top part of a random abstract ASCII tree.

We are testing our **MOI** on their general tree capabilities: can they understand and navigate abstract trees of increasing size? For instance, the largest random tree used in this experiment has a thousand nodes and a maximum depth of nine. Can LLMs handle these?

For our experiment, we ask the following questions of each model for each tree size. The actual prompts (Appendix D) are more detailed but the following table gives the reader an idea of the nature of the questions used in this experiment. These questions, and the wide range of tree sizes, were designed to push the LLMs to their point of failure. Each tree size had random arrangements of parents and children, and so the three question marks in these questions varied with the tree size.

| Ask ID | The Ask |
|---|---|
| **T1** | How many nodes are in the tree above? |
| **T2** | Build a complete ASCII tree visualization of the hierarchy in the tree above'. Use standard tree-drawing rules: use ├── for branches, └── for last child, etc. |
| **T3** | Give a JSON artifact with the direct children of the node named '???' from the tree above.. No grandchildren. No parents. No siblings. |
| **T4** | Give a JSON artifact with the UPTREE and DOWNTREE nodes from the node named '???' in the tree above. |
| **T5** | Determine the DOWNTREE nodes from the node named '???' in the tree above. |

Table 5: Theory questions used in the first experiment.

| n | LLM Model | Took (secs) | Ave Grade | Std Dev | *Cost (cents) |
|---|---|---|---|---|---|
| 25 | gemini-2.5-pro | 55.4 | 98.8 | 2.25 | 2.35 |
| 25 | claude-opus-4-5 | 19.8 | 97.7 | 5.13 | 5.88 |
| 25 | grok-code-fast-1 | 24.6 | 92.7 | 15.2 | 0.23 |
| 25 | gemini-3-pro-preview | 77.2 | 91.3 | 27.5 | 2.44 |
| 25 | grok-4-1-fast-reasoning | 43.7 | 87.7 | 28.8 | 0.39 |
| 25 | claude-sonnet-4-5 | 23.0 | 87.1 | 24.6 | 3.65 |
| 25 | gpt-5.2 | 0.01 | 84.2 | 23.3 | 2.61 |
| 25 | gemini-2.5-flash | 52.4 | 78.6 | 40.5 | 0.44 |
| 25 | claude-haiku-4-5 | 9.52 | 74.1 | 36.3 | 1.18 |
| 25 | gpt-5.1 | 0.01 | 69.5 | 35.1 | 1.44 |
| 25 | grok-4-1-fast-non-reasoning | 5.03 | 64.1 | 36.5 | 0.21 |
| 25 | gpt-4.1 | 20.0 | 57.2 | 38.5 | 1.61 |

Table 6. Experiment 1: Testing 12 models on general tree capabilities. 25 tree theory questions.

Above are the results from experiment 1. All LLMs were given the same single shot prompt, issued against the model vendor's API and five sizes of random abstract trees.

Each question submitted to the MOI has a known right answer. Partial credit was given for close answers. For example, if the correct answer to question 1 was 50 but the LLM responded with 49, then the model would be scored with a 98.

The 'Took' column was calculated by taking the difference (in seconds) from the moment the shot was issued to the LLM until the moment the response was received.

*Cost column: Token counts used in the calculation of the cost column are returned from each vendor in their API response JSON ('tokens in' and 'tokens out'). The Cost is then computed per the published rates (as of Dec 15th 2025) on each vendor's model pricing sheets. See Appendix E for more details.

The results shown are hopeful: some models exhibit a high degree of understanding of abstract trees. However, the results also show there is a fair degree of variation in MOI's ability to work with abstract trees, from 57% to 99%. Later we will see that this 'abstract tree' ability does not predict the ability of a model's effectiveness in the real world use case of auto-coding tasks.

## 8 LLM Auto-Coding Performance with TREEFRAG (LOD 1)

According to table 6, our MOI understand trees to some degree. How well will this ability transfer to general knowledge and tree capabilities in our real world use case of auto-coding?

### 8.1 Experiment 2A

To address this question, we evaluated all 12 of our Frontier LLM MOI on 40 real-world auto-coding tasks (bugs, enhancements, refactors) using **TREEFRAG** at LOD 1 (node names only, no code bodies or summaries). The prompt contained the **RWT** tree skeleton plus the task description. The task descriptions are bugs, enhancement and refactor requests as found in Appendix C. This experiment evaluated issue location / understanding only; experience shows that precise scoping ensures reliable code fixes.

Table 7 below shows results. **TREEFRAG** compresses the **RWT** raw codebase (194k to 239k) to ~10k tokens (18:1 to 24:1 ratio, model-dependent). All models achieved 94.5--97.3% success, with grok-4-1-fast-non-reasoning leading (97.3% at 0.20¢ per task).

These results demonstrate that LLMs can perform high-quality auto-coding tasks from hierarchical metadata alone, validating **TREEFRAG'**s core claim: structure + concise node names convey sufficient architectural information for precise task execution.

| n | LLM Model | Code Base (tokens) | TREEFRAG (tokens) | Comp Ratio | Took (secs) | Ave Grade | Std Dev | Cost (cents) |
|---|---|---|---|---|---|---|---|---|
| 40 | grok-4-1-fast-non-reasoning | 195,451 | 8,333 | 23:1 | 5.20 | 97.30 | 2.11 | 0.20 |
| 40 | claude-haiku-4-5 | 210,519 | 11,474 | 18:1 | 9.10 | 96.70 | 2.05 | 1.82 |
| 40 | gpt-5.2 | 196,361 | 8,255 | 24:1 | 0.01 | 96.40 | 1.59 | 3.44 |
| 40 | claude-sonnet-4-5 | 210,519 | 11,474 | 18:1 | 17.52 | 96.40 | 1.78 | 5.13 |
| 40 | grok-4-1-fast-reasoning | 195,451 | 8,239 | 23:1 | 16.92 | 96.30 | 1.93 | 0.20 |
| 40 | claude-opus-4-5 | 210,519 | 11,470 | 18:1 | 16.63 | 96.10 | 1.80 | 8.61 |
| 40 | gemini-2.5-pro | 239,154 | 12,139 | 20:1 | 32.54 | 95.60 | 1.64 | 2.79 |
| 40 | grok-code-fast-1 | 195,451 | 8,375 | 23:1 | 18.22 | 95.40 | 2.77 | 0.28 |
| 40 | gpt-4.1 | 196,361 | 8,255 | 24:1 | 23.27 | 94.90 | 2.85 | 2.61 |
| 40 | gemini-2.5-flash | 239,154 | 12,140 | 20:1 | 23.55 | 94.90 | 1.85 | 0.77 |
| 40 | gpt-5.1 | 196,361 | 8,255 | 24:1 | 0.01 | 94.70 | 1.56 | 2.51 |
| 40 | gemini-3-pro-preview | 239,154 | 12,135 | 20:1 | 29.45 | 94.50 | 2.82 | 3.64 |

Table 7. Results from 12 models across 40 coding tasks showing compression ratios, average grades and cost per prompt.

Grading was by consensus: all members of the MOI grading each other member of the MOI. Consensus grading follows common practice in LLM evaluation to average model judgments. All models were given the same prompt. The prompt used for this experiment can be found in Appendix D: basically the MOI were tasked with returning the tree nodes that the model thought were most related to the users 'Ask' (bug, enhancement, etc.) along with their estimate of relatedness expressed as a probability. From all answers given by all models a 'grading key' was created, but if a tree node was only reported by one model it was eliminated from the key. Grades were then determined by how closely a given model's answers matched the consensus key.

Oddly, Table 7's coding performance rankings somewhat invert Experiment 1's abstract tree results (Table 6): For instance, Gemini leads in working with abstract trees but trails in real tasks; Grok and Claude dominate production code taskings. This highlights **TREEFRAG'**s strength: real-world software hierarchies (GUIs, DBs, specs) unlock models' true potential, favoring code-trained LLMs like Claude and Grok over generalists.

However, in Table 7 we see all MOI scored above 94%, showing our MOI LLMs are generally quite competent at our test taskings using the **TREEFRAG** exploit**.**

But how well does **TREEFRAG** compare to other methods of auto-coding?

### 8.2 Experiment 2B: Grading Study on Issue-Resolution Specifications

To evaluate whether **TREEFRAG** truly produces superior issue-resolution specifications compared to baseline methods, we conducted a large-scale grading study. For each of 40 real-world issue reports (Asks), we have our MOI generate 45 specifications (45 total per Ask, ~1,800 total) using four methods:

1. **Naive** method: The raw code base is sent, along with the Ask. This is called the naive method as it's simple to construct the context prompt. This method is likely every dev's first 'auto-coding' experiment. In this method, the **RWT's** codebase is sent 'raw', so tokens vary between 194k to 239k depending upon the LLM model, producing no compression (1:1) advantage.

2. **File Summary** method: In this method, each file in the code base is summarized by an LLM, reducing the token count significantly. Details are lost but these may not be needed to identify the location of the Ask's issue or understand its impact in the code base. This is equivalent to **RWT** LOD 6 on file nodes only: tokens are 20k to 23k depending upon the LLM model. Compression ratios are between 9:1 and 10:1.

3. **Function Summary** method: This method is more sophisticated and requires database support to properly implement. Token counts balloon over the file summary method 2 above because each function in the code base is summarized individually. We believe this method is the state of the art for auto-coding currently. Token counts are 53k to 61k depending upon the LLM model. Compression ratios are between 3:1 and 4:1.

4. **TREEFRAG** method: In this method, only the **RWT** hierarchical tree is sent at LOD 1 (node names only) along with the user's Ask. Token counts are between 8k and 12k. Compression ratios are between 18:1 and 24:1 depending upon the LLM model.

All specifications were graded by the top model (grok-4-1-fast-non-reasoning in table 7) using the task prompt in Appendix D. 1800 shots: 40 tasks (Asks) across 12 models and 4 methods minus the Claude models for method 1 (Naive) shots (due to context window size limitations).

To minimize potential bias by the grading LLM, the model being graded was hidden: we blinded the grader by anonymizing all identifiers and shuffling report order within batches, preventing model identification: grader was given a random selection of 5 issue reports per shot. There were nine such groups of 5 reports per Ask.

**Method: Mean Scores**

Below are the results of Experiment 2B: Blinded Grading of Issue-Resolution Specifications.
Mean scores across all tasks were tight:

| Method | n | Mean Score | Std Dev |
|---|---|---|---|
| Function Summary | 450 | 92.05 | 4.72 |
| TREEFRAG | 450 | 91.95 | 4.48 |
| Naive | 450 | 91.24 | 6.70 |
| File Summary | 450 | 90.98 | 5.77 |

Table 8: Blinded Grading Experiment 2B: Mean Scores Across Methods

Mean scores were tightly clustered, differing by about 1 point overall. This narrow spread and standard deviations greater than 4 implies that average scores alone are insufficient for declaring a method as a definitive winner, necessitating deeper per-task analysis via ranking.

**Method: Mean Rank**

We therefore computed per-Ask ranks (1 best, 45 worst). **TREEFRAG** showed the lowest mean rank (best performance):

| Method | n | Mean Rank | Std Dev | First Place |
|---|---|---|---|---|
| TREEFRAG | 450 | 20.69 | 3.38 | 22 / 40 |
| Function Summary | 450 | 21.34 | 3.01 | 11 / 40 |
| Naive | 450 | 24.69 | 4.82 | 5 / 40 |
| File Summary | 450 | 25.70 | 3.37 | 2 / 40 |

Table 9: Experiment 2B Per-Task Mean Ranks Across Methods (lower rank = better)

Expected random rank: 23. Monte Carlo simulation (1,000 trials) confirmed a random mean of 22.99 ± 0.51. **TREEFRAG's** rank mean of 20.69 is about 4.5σ standard deviations below random, proving consistent outperformance. Although method mean scores were close, per-Ask ranking reveals **TREEFRAG's** superiority: mean rank 20.69 vs random 23 (4.5σ, $p \ll 0.0001$), and first place in 22 of 40 Asks.

Ranking reveals that **TREEFRAG** produces the best User Issue Specification Report most often, even from a skeleton of node names alone. This supports our thesis: hierarchical structure enables superior issue analysis and specification quality, even with token compression rates of 18:1 to 24:1 over raw code base tokens.

## 9 Discussion

The results validate **TREEFRAG**'s core hypothesis: hierarchical homogenization and metadata-only representation (LOD 1) achieve 18:1 to 24:1 token compression while enabling frontier LLMs to locate and understand real-world issues with 94 to 97% accuracy at cents-per-task cost. Grok models consistently lead in both performance and economics - often 10 to 20x cheaper than competitors.

Experiment 2B's blinded ranking study reveals **TREEFRAG**'s edge: despite near-identical mean scores, it ranks first in 22 / 40 tasks (vs. expected ~9 under random), a 4.5σ deviation confirming consistent superiority in issue specification quality. We believe this stems from **TREEFRAG**'s ability to preserve architectural relationships that flat methods discard.

Operationally, **TREEFRAG** excels on economics: Grok delivers equivalent accuracy at the lowest cost and near highest compressibility. For users prioritizing sub-second latency on large prompts, OpenAI models remain very fast but at higher prices.

Caveats remain:

- **TREEFRAG** adds implementation complexity; simple apps (<1k lines) gain little.
- **TREEFRAG** requires the creation of a tree, requiring a decomposition pre-process.
- Token pricing volatility (20x spread across vendors) threatens cost models: with such a wide price spread, current inference pricing at the low end may not be sustainable.
- Hard context limits (e.g., Claude's strict 200k cutoff) can silently break workflows.

- LLM-generated node names/tag_lines may embed latent cues aiding comprehension - though an informal experiment swapping tag_lines for node_names yielded identical outcomes, suggesting structural rather than lexical magic.

These results reveal several key insights. First, Stingy Context achieves over 90% token compression while retaining high task fidelity on real codebases, demonstrating that most prompt tokens are low-utility for issue localization. Second, the compressed **TREEFRAG** representation is more human-readable than raw source code—transforming 1D linear text into a structured, navigable 2D hierarchy. Third, frontier LLMs reliably interpret these reduced prompts, suggesting tree-native reasoning is broadly accessible. Fourth, the compression yields substantial inference cost savings (often 10–20× cheaper than baselines). The exploit is conceptually simple, obvious in hindsight, and medium effort to implement, yet it addresses a core limitation in current spec-driven auto-coding: *conveying software architecture efficiently to LLMs*. While flat methods remain common, **TREEFRAG** outperforms them in both cost and quality on production tasks.

Scaling laws favor ever-larger contexts, but **TREEFRAG** shows compression is the higher-leverage path: logarithmic token growth versus linear raw code tokens. For production-grade auto-coding, hierarchical trees appear indispensable.

## 10 Related Work

Long-context extensions like **LongRoPE** (Microsoft, 2024) retrofit RoPE embeddings to reach 2M+ tokens but require model-specific retraining and scale compute quadratically. **TREEFRAG** achieves similar effective scale via preemptive compression without modifying models.

RAG-based coding tools (e.g., **Retrieve-and-Generate** in Cursor, GitHub **Copilot**) chunk code and embed for retrieval, but flatten hierarchy, losing dependencies and suffering relevance drift. **TREEFRAG** preserves full structure in a homogenized tree.

File/function summarization tools (e.g., **Aider**'s **Repo Map**, **Continue.dev**'s context providers) reduce tokens via hierarchical or graph-based abstracts and apply task-aware ranking/retrieval. However, these remain post-hoc overlays on linear source code and lack **TREEFRAG**'s unified, multi-domain homogenization (code + GUI + DB + specs) into a single traversable tree. **TREEFRAG** enables accurate reasoning from metadata skeletons alone, outperforming even dynamic graph methods in blinded ranking (first place in 22 of 40 tasks).

**FrugalPrompt** (2025) is a very generalized approach to context window compression, working on the context window tokens across domains. It compresses prompts by saliency-based token retention, achieving ~20% reduction on NLP tasks with minimal degradation, but fails on mathematical reasoning where full context continuity is critical. Unlike **TREEFRAG**'s structural hierarchy preservation, **FrugalPrompt** operates on flat sequences and lacks multi-domain homogenization or LOD control.

| Method | Compression Ratio | Hierarchy Preserved | Multi-Domain (Code+GUI+DB+Specs) | LOD 1 Viable | Setup Effort |
|---|---|---|---|---|---|
| Naive Chunking | 1:1 | No | No | No | Low |
| LongRoPE | N/A (extends window) | Yes (positional) | No | N/A | High (retrain) |
| Repo Maps Aider | 3:1–10:1 | Partial | No | No | Medium |
| FrugalPrompt | 2:1 | No | No | No | Medium |
| GraphRAG | 5:1–15:1 | Partial | Partial | No | High |
| ASTs | 5:1 - 20:1 | Full (syntax) | No | Partial | High |
| Call Graphs | 10:1 - 50:1 | Partial | No | No | High |
| RAGFlow | 2:1 or 5:1 | No | No | No | Medium |
| TREEFRAG | 18:1–24:1 | Full | Yes | Yes | Low/Medium |

Table 10. Related works to the **TREEFRAG** exploit

Hierarchical graph techniques for knowledge extraction (e.g., **GraphRAG**, Microsoft 2025, building on earlier work such as US Patent 10,528,668, (2020) target document QA. **TREEFRAG** adapts and extends this paradigm to software architectures, unifying code, GUI, database, and specifications into a single homogenized tree, purpose-built for auto-coding.

Prior tree representations (**AST**s, call graphs) aid static analysis but lack LLM-optimized serialization or multi-domain homogenization. **TREEFRAG** advances these by enabling accurate auto-coding from metadata skeletons alone.

**RAGFlow** (2024) provides open-source RAG pipelines with document chunking and retrieval, but operates on flat text without **TREEFRAG**'s multi-domain hierarchical homogenization or LOD 1 metadata compression.

**Stingy Context** complements rather than competes with long-context scaling: compression makes large windows feasible and affordable for iterative development.

## 11 Conclusion

Stingy Context delivers what brute-force long contexts cannot: 18:1–24:1 token compression with 94–97% issue-location accuracy on a real 20,756-line codebase, at cents per task.

**TREEFRAG** shows hierarchical homogenization transforms flat code into fractal-like structures LLMs navigate with uncanny precision—even from node names alone.

The era of raw-code prompts and naive chunking is over. Hierarchical compression is no longer optional; it is the only viable path forward. Grok models already dominate these benchmarks at less than one cent per task; the labs embracing tree-native methods first will own the next decade of software development.

This paper has focused on **TREEFRAG**, the foundation of Stingy Context. Ahead lies Variable Focus and Blur (**VFAB**) for task-specific sharpening and survey-driven dynamic specification capture.

We release Stingy Context into the wild—implement it, extend it, beat it.

## Bibliographical References

## Appendix A RWT: Detailed Description of the ViperPrompt application

The ViperPrompt application, referred to as **RWT** in this paper:

- Is 100% LLM-auto-coded (except Mathematica compilers). LLMs that were used to auto-code ViperPrompt include OpenAI, Anthropic and xAI models over the course of the last 7 months.
- Tree node names were automatically generated by the auto-coding models as needed.
- Prior versions of ViperPrompt were used to code itself in a bootstrapping fashion.
- There are 1,014 nodes in the current **RWT** tree used in preparation of this paper.
- Currently, ViperPrompt has 8 code bases across three languages: Python, RUST and Mathematica, plus some miscellaneous BAT files.
- There are 20,756 lines of source code in total among all the code bases.

| Name | Type | Auto Coded | Language | Lines | Description |
|---|---|---|---|---|---|
| ViperTree | GUI | 100% | Python PySide6 | 7181 | A GUI allowing the advanced ViperPrompt user to modify their ViperPrompt tree. Users can construct their own superstructures around their code base(s) to support different software development methods such as Scrum, Agile, Waterfall, etc. |
| ViperVSL | CLI | 100% | Python | 2350 | VSL runtime engine. Actively being translated from Mathematica. |
| ViperAPI | CLI | 100% | RUST | 2821 | Manages API calls to LLM vendors. |
| ViperAuto | CLI | 100% | Python | 847 | Automatically maintains the ViperPrompt tree according to the user's vision. Runs automagically and employes multiple LLMs in an agentic fashion. |
| ViperThread | GUI | 100% | Python PySide6 | 1433 | Interfaces with the user, collecting their vision and specs for their application. |
| ViperPrompt Meta Compiler | .NB | N | Mathematica<br><br>Meta language | 449 | Single pass, recursive descent compiler compiler, with Backus-Naur Form (BNF) grammar definitions which can be interlaced with object code production rules. May be translated to RUST. |
| VSL Compiler | .NB | N | Mathematica<br><br>Meta language | 2620 | Compiles code written in VSL (ViperPrompt Scripting Language). VSL is a purpose built language allowing developers to easily and clearly embed SQL and logic with prompt text. Much of ViperPrompt is written in VSL so users can modify ViperPrompt for their own purposes. Being translated to Python. |
| VSL Runtime | .NB | 50% | Mathematic a | 3055 | Stop gap VSL runtime engine. Actively being translated to Python. See ViperVSL above. |

## Appendix B Full Category-Type Table

Below is the contents of the category-type table used by the **RWT**.  This table is user-extendable for different use cases.

| Category | Type | Description |
|---|---|---|
| Structure | Scafold | Tree node - scaffold structuring |
| Process | Project | Project |
| Process | Status | Process status update |
| Process | Milestone | Project milestone |
| Process | Benchmark | Performance benchmark |
| Process | Version | Version... like V1.1 |
| Process | SetMark | Process marker |
| Process | Lesson Learned | Lesson learned note |
| Process | Note to Self | Personal note |
| Process | Note Going Forward | Forward-looking note |
| Process | Work Product | Process work product |
| Process | Notification | Process notification |
| Process | Backup | Backup description |
| Process | FormalLog | Formal process log |
| Specification | Vision | Vision statement for project direction |
| Specification | Concept | High-level project concept |
| Specification | Idea | Initial project idea or proposal |
| Specification | Spec | Detailed specification document |
| Specification | Feature | Specific feature description |
| Specification | Requirement | Project requirement or need |
| Specification | Standard | Defined standard or guideline |
| Specification | Risks | Description of Risk(s) |
| Specification | Option | Optional feature or choice |
| Specification | Issue | Identified issue or concern |
| Specification | Comment | General comment or note |
| Specification | Acronym | An acronym of interest |
| Standard | RUST | RUST coding standards |
| Standard | Python | Python coding standards |
| Standard | Template | Standards for Templates |
| Patent | Folder | Patent Folder holds patent files |
| Patent | Provisional | Provisional Patent PDF |
| Patent | Non Provisional | Non Provisional Patent PDF |
| IP | Patents | Patent Folder holds patent files |
| Code | Function | Usually a child of a code node |
| Code | RUST | Rust source code |
| Code | Python | Python source code |
| Code | BAT | Batch script |
| Code | PowerShell | PowerShell script |
| Code | Folder | Folder for code |
| Code | Executable | Executable |
| Code | Template | A Templex template |
| Code | Meta | Meta code |
| Code | Mathematica | Mathematica code |
| Code | PySide6 | Extension of *.ui describes PySide6 |
| Generic UI | Button | UI button component |
| Generic UI | Text Input | Single-line text input |
| Generic UI | Label | UI label component |
| Generic UI | Message | UI message display |
| Generic UI | Dialog | UI dialog box |
| Generic UI | Text Multi-line | Multi-line text input |
| Generic UI | ComboBox | UI dropdown menu |
| Generic UI | Panel | UI panel container |
| Generic UI | Grid | UI grid layout |
| Generic UI | Checkbox | UI checkbox component |
| Generic UI | Tree | UI tree view |
| Generic UI | Table | UI table component |
| PySide6 | Main Window | Application window with menu, toolbar |
| PySide6 | Widget | Base class for UI elements |
| PySide6 | Horizontal Layout | Arranges widgets horizontally |
| PySide6 | Vertical Layout | Arranges widgets vertically |
| PySide6 | Grid Layout | Arranges widgets in a grid |
| PySide6 | Form Layout | Arranges widgets in a two-column form |
| PySide6 | Frame | Container with frame styling |
| PySide6 | Group Box | Container for grouping widgets |
| PySide6 | Dock Widget | Movable, dockable widget |
| PySide6 | Tab Widget | Tabbed widget container |
| PySide6 | Scroll Area | Scrollable widget area |
| PySide6 | Splitter | Resizable widget splitter |
| PySide6 | Menu | Context or dropdown menu |
| PySide6 | Tool Bar | Toolbar with actions |
| PySide6 | Status Bar | Displays status information |
| PySide6 | Graphics View | Scene-based graphics display |
| PySide6 | List View | List-based item display |
| PySide6 | Tree View | Hierarchical tree display |
| PySide6 | Table View | Spreadsheet-like table display |
| PySide6 | Table Widget | Pre defined columns |
| PySide6 | Line Edit | Single-line text input |
| PySide6 | Text Edit | Multi-line text input |
| PySide6 | Plain Text Edit | Lightweight multi-line text editor |
| PySide6 | Combo Box | Dropdown selection menu |
| PySide6 | Check Box | Toggleable checkbox |
| PySide6 | Radio Button | Exclusive radio button |
| PySide6 | Slider | Value slider control |
| PySide6 | Dial | Circular control for selecting values |
| PySide6 | Spin Box | Numeric input with spin controls |
| PySide6 | Double Spin Box | Floating-point numeric input |
| PySide6 | Date Edit | Date-only input |
| PySide6 | Time Edit | Time-only input |
| PySide6 | Date Time Edit | Date and time input |
| PySide6 | Calendar Widget | Date selection calendar |
| PySide6 | LCD Number | Digital display for numbers |
| PySide6 | Progress Bar | Progress indicator |
| PySide6 | Tool Button | Button for toolbar or quick actions |
| PySide6 | Button | Standard push button |
| PySide6 | Label | Text or image display |
| SQLite | Database | Name of the SQLIte database |
| SQLite | Text field | SQLite TEXT field |
| SQLite | Integer field | SQLite INTEGER field |
| SQLite | DateTime field | SQLite Timestamp field |
| SQLite | Real field | SQLite Real field |
| SQLite | Table | SQLite TABLE |
| SQLite | Text Value | SQLite enumerated text value |
| SQLite | Number Value | SQLite enumerated number value |
| SQLite | Calculated Number | SQLite calculated number field value |
| SQLite | Calculated Text | SQLite calculated text field value |
| Database | Text field | SQLite text field |
| Database | Table | SQLite table |
| Database | Database | Generic Database |
| Database | SQLite | SQLite database |
| Database | PostGres | PostGres database |
| Ask | Bug | Reported bug |
| Ask | Issue | Testing issue |
| Ask | Compile Error | Testing error |
| Ask | Refactor | Testing protocol |
| Ask | Instructions | Testing instructions |
| Ask | Assessment | Testing assessment |
| Ask | Enhancement | Test run description |
| Ask | Plan | Overall project plan |
| Ask | Step | Individual step in a plan |
| Paper | Abstract | Paper abstract |
| Paper | Executive Summary | High-level summary |
| Paper | Body | Main content of paper |
| Paper | Summary | Concluding summary |
| Paper | Conclusion | Paper conclusion |
| Paper | Index | Index of paper sections |
| Paper | References | List of references |
| Paper | Section | Paper section |
| Paper | SubSection | Paper subsection |
| Paper | SubSubSection | Paper sub-subsection |
| Book | Prolog | Book prologue |
| Book | Forward | Book foreword |
| Book | Acknowledgements | Book acknowledgments |
| Book | Table of Contents | Book table of contents |
| Book | Chapter | Book chapter |
| Book | Appendix | Book appendix |
| Book | Index | Book index |

## Appendix C Task Table

Below are the 40 'Asks' used in experiments 2A and 2B. Many of these were actual bug, enhancement, and refactor requests encountered during the auto-coding of the **RWT** application.

| Type | ID | Ask |
|---|---|---|
| Bug | B1 | After I move a node with drag-and-drop in ViperTree, the project_id for some of its grandchildren is not updated and stays pointing to the old project. This breaks DOWNTREE(project) selections later. |
| Bug | B2 | When I paste a copied subtree that contains code nodes, the node_hash and compile_hash fields stay empty forever on the pasted nodes, so ViperAuto never detects changes and never re-imports the artifacts. |
| Bug | B3 | ViperVSL COUNT directive does not work inside an IF block. It always returns 0 even when the preceding SELECT clearly has rows. |
| Bug | B4 | When ViperAPI receives a response with multiple artifacts of the same type (e.g. two Python files), only the last one gets written to disk. |
| Bug | B5 | When I use VSL directive SAVESELECT(mine) and later LOADSELECT(mine) inside the same template, the loaded selection is always empty on the second and subsequent generations. |
| Bug | B6 | ViperAuto is importing the same artifact twice when two different shots produce a file with the exact same name and content (creates utils.py etc.). |
| Bug | B7 | The FILE-CONTENT(node_id) directive throws "path not found" if the where_path ends with a trailing backslash on Windows. |
| Bug | B8 | The temperature setting in the shot table is not set according to the CLI. Every shot uses the default from config.py. |
| Bug | B9 | Node MOVE UP and DOWN sometimes wrong after copy-paste. |
| Bug | B10 | When I click "Sync Children" on a folder node that contains subfolders, ViperTree only imports first-level files and ignores deeper levels. |
| Bug | B11 | The ASCII tree dump is cutting off node_names longer than ~60 characters. |
| Bug | B12 | If INTERSECTION(mykey) and the saved SELECT is empty, ViperVSL crashes with "index out of range". |
| Bug | B13 | ViperAuto sometimes creates a new shot record with shot_type = "autotest" but leaves prompt and Viperprompt columns NULL, causing deadlock. |
| Bug | B14 | When I use WHERE(read) on a file that has $$ directives inside, ViperTemplex expands them too early and breaks the file. |
| Bug | B15 | After a long session (>200 shots), ViperThread's "Next/Previous" conversation buttons start skipping shots randomly. |
| Bug | B16 | The 'what' column is sometimes filled with the wrong information, causing downstream artifact processing problems. |
| Bug | B17 | It doesn't work |
| Bug | B18 | The ViperAPI processes shot records via API calls to Lab models. Sometimes it hangs forever waiting for a response. Limit to 5 minutes |
| Bug | B19 | I can apparently make a backup of the Viper system OK but when I try to restore from a backup it may or may not work, causing me to have to restore further and further back in time to find a good restore. |
| Bug | B20 | When I try to set a folder for a where path it sometimes does not work and does not sync |
| ENH | E1 | Can we please get a "Reveal in Explorer / Open Folder button on any node that has a where_path filled in? Right-click \[RightArrow] "Open folder" would save a ton of time instead of copying the path manually. |
| ENH | E2 | Add a simple "Regenerate this node" right-click option on any node (uses the same template/rules that created it the last time, with current context). 90% of my iterations are just "try again with slightly better focus". |
| ENH | E3 | Add a dark mode / proper Qt style sheet support to ViperTree and ViperThread. |
| ENH | E4 | In ViperThread, add a small "Attach current runtime.log / compiler.log / console.log" checkbox trio directly under the Ask text box. |
| ENH | E5 | Please add "Find in Tree" (Ctrl+F) to ViperTree that searches node_name, tag_line, short_desc AND content at the same time. |
| ENH | E6 | Can we have a right-click \[RightArrow] "Promote to Project Root" on any node? |
| ENH | E7 | Add a tiny generation counter badge (e.g. "Gen 42") next to the project name in the ViperTree title bar. |
| ENH | E8 | ViperThread needs a "Copy full prompt to clipboard" button after a shot finishes. |
| ENH | E9 | Please add undo/redo (at least 20 levels) for node create/move/delete/copy/paste in ViperTree. |
| ENH | E10 | In ViperThread, add token counter breakdown (NOI+VFAB, conversation history, logs, ask). |
| ENH | E11 | Right-click on any code node \[RightArrow] "Open in VS Code" that passes the full where_path + filename. |
| ENH | E12 | Add a "Prune ignored/future/deprecated subtrees from context" checkbox in ViperThread. |
| ENH | E13 | Add a tiny "pin" icon next to important nodes so they always stay visible at the top of the tree. |
| ENH | E14 | Add a one-click "Mark entire subtree as deprecated/ignored" right-click option. |
| ENH | E15 | The ViperAPI processes shot records via API calls to Lab models. But the console only shows the time hack and the cost. Add the model called. |
| REF | R1 | The tree widget in ViperTree is getting slow with >3000 nodes. Please add lazy loading or virtual mode for the QTreeWidget so only visible branches are loaded into memory. Even just collapsing a big project makes the GUI freeze for 3 to 4 seconds. |
| REF | R2 | The backup system folder names are just timestamps. Please add an optional description field or use root node name + generation number. |
| REF | R3 | Go through all the source code base and regenerate each log line so the log tag ID is based on the sub project AKA field. |
| REF | R4 | What is involved in having a config-type switch that tells ViperPrompt to use SQLITE or Postgres as the underlying database? |
| REF | R5 | Current ViperAPI behavior marks shots as failed when API errors occur (404, timeouts, etc.). This enhancement proposes leaving such shots in a locked state until manual intervention or retry logic is implemented. Benefits include better error tracking, preventing downstream processing of incomplete data, and improved debugging of API connectivity issues. |

## Appendix D Reproducibility: Example Prompts

Below are 3 prompts that were used in the production of this paper. These reproducibility prompts are redacted to protect ViperPrompt's intellectual property. Full **TREEFRAG** dumps and source code from the **RWT** codebase are withheld to prevent direct replication of patent-pending structures and algorithms, while still demonstrating exact prompt formats and methodology for independent verification of claims.

These prompts have been reduced in size: cut and paste to examine them in more detail.

### Redacted prompt used in the production of Experiment 1, Table 6:

````text
Consider the following dataset, which we will name 'T':

"id","parent_id","node_name"
15991,0,"a"
1906,15991,"aa"
90,15991,"ab"
1627,1906,"aaa"
2078,15991,"ac"
14736,1627,"aaaa"
2638,14736,"aaaaa"
14117,15991,"ad"

Etc.

You are a very good Computer Science student named Alice.

TASK:  Answer this question: how many nodes are in dataset 'T'?

This is Question ID: 'T_1_XSmall_1'.

Only respond with the correct answer in the JSON artifact format below.  Do NOT include any explanation of any sort.  There should be nothing in your response except the artifact with your answer.

Your answer should follow this JSON artifact format:

--- Artifact Start ---
File: ask-T_1_XSmall_1.json
```json
{
  "question_id": "T_1_XSmall_1",
  "number_of_nodes": 22
}
```
--- Artifact End ---

 Do not forget: your artifact should have Start and End indicators as shown in the example above.  Please check your work carefully before responding.
````

## Redacted prompt used in the production of Experiment 2A, Table 7:

~~~~
The following is a tree structure which describes the work-in-progress ViperPrompt Application.

  ViperPrompt(18)
     Vision(23)
        The Big Picture(256)
        ViperPrompt high level(3392)
        Features(259)
           No Code - Low Code(265)
Etc.
--------------------------------------------------------------------------
Instructions:

Below is an issue report by a user of the ViperPrompt system.  In their Issue Report a user says the following:

'After I move a node with drag-and-drop in ViperTree, the project_id for some of its grandchildren is not updated and stays pointing to the old project. This breaks DOWNTREE(project) selections later.'

---------------------------------------------------------------------------

TASK: We want to resolve the user's issue.  To do this, we first want to isolate what parts of the system are involved with the issue.  Please give a short list of the 20 top nodes that are most related to the user's issue.  There may be just a few or one node of interest.

Be sure to consider these, which are in order of importance:

1.  The node of the root 'uber' project involved in this issue.
2.  The project node where this issue occurs in the application.
3.  Node(s) of any source code that may need changes to address the user's issue.
4.  Node(s) of any source code that should be considered.
5.  Node(s) of any data base table that is involved in this issue.
6.  Node(s) of any GUI that may be affected by this issue.
7.  Other node(s) of interest to address the user's issue.

Use this format below to indicate the node(s) of interest:

Artifact JSON format

All LIKELIHOOD (probability) assessments should produce an artifact like the example below.  Notice that it has a header 'File: NodeProbability.json' followed by a single Markdown code block using the language tag matching the file extension: ```json.

Match the following example JSON format exactly.   Also, all artifacts are to be wrapped with start and end indicators as shown in this example:

--- Artifact Start ---
File: ask-B_1_Viper_1-NodeProbability.json
```json
[
  {
    "node" :18,
    "node_name": "ViperPrompt",
    "category": "Process",
    "type": "Project",
    "reason": "This node is the root for subject matter around the user's topic of interest",
    "probability": 90
  },
  {
    "node" :2872,
    "node_name": "context menu mnager.py",
    "category": "Code",
    "type": "Python",
    "reason": "Manages the ViperTree context menu",
   "probability": 65
  }
]
```
--- Artifact End ---

The probabilities are expressed as percentages and need NOT add up to 100 percent:  only a small number of node IDs out of the complete set of nodes may be given.

Also, you may be asked to select up to 20 top probabilities, but a much smaller number may suffice, in which case the number of node IDs in your JSON will be fewer than requested.  It will be assumed that any node IDs not included in your JSON have probabilities close to zero.

Remember that all artifacts are to be wrapped with start and end indicators as shown in the example above.  Do NOT return any other information besindes the requested probability JSON described above.  Be sure to name your JSON file:

File: ask-B_1_Viper_1-NodeProbability.json
~~~~

# Redacted prompt used in the production of Experiment 2B, Tables 8 and 9:

````
RWT source code base inserted here:  LOD 7.

RWT file and function summaries inserted here:  LOD 6.

RWT tree inserted here.

------------------------------------------------------------------------
User Issue Report

A user has reported the following issue in the above described system:

After I move a node with drag-and-drop in ViperTree, the project_id for some of its grandchildren is not updated and stays pointing to the old project. This breaks DOWNTREE(project) selections later.

---------------------------------------------------------------------------------

Five students have turned in their Spec reports and they are given below.

-----------------------------------------------------------------------------
SpecID = B_1_1_O_51

Spec Text follows:

Issue Summary
After a node is moved via drag-and-drop in ViperTree, some of its grandchildren retain the old project_id, so later DOWNTREE(project) queries operate on an incorrect project scope.

Root Cause Hypothesis
Drag-and-drop reparenting is handled by QTreeView's built-in move behavior, not by the explicit move_node logic in node_operations.py. The explicit move_node path correctly calls update_project_id_downtree(node_id) after a move, but the drag-and-drop path never invokes this function and likely does not update project_id in the database for the moved subtree. As a result, only some nodes (e.g., those touched by other operations) get corrected project_id values, while deeper descendants keep their original project_id, causing DOWNTREE(project) and related directives to see a mixed project tree.

Components of Interest
- main.py (HyperTree GUI entry point; QTreeView configuration via Ui_MainWindow)
- tree_manager.py
  - TreeManager.setup_tree (tree view configuration, drag-and-drop settings)
  - TreeManager (place to hook model move events)
- node_operations.py
  - NodeOperations.move_node
  - NodeOperations.update_project_id_downtree
- panel_manager.py
  - PanelManager.update_project_ids_for_node (Process/Project changes)
- sync_manager.py
  - SyncManager.on_sync_clicked / process_directory (uses project_id for Folder/DOWNTREE)
- database schema: tree table (project_id, parent_id, root_id, depth)
- directive_processing_logic.py / directive_handlers.py
  - DOWNTREE(project) and related directives that rely on project_id

Proposed Changes
- tree_manager.py
  - In setup_tree, explicitly configure the QTreeView/model to disable internal drag-and-drop moves, or set dragDropMode to NoDragDrop if drag-and-drop is not required.
  - If drag-and-drop must remain, subclass QStandardItemModel or intercept the model's rowsMoved signal in TreeManager to detect when a node is reparented. In the rowsMoved handler:
    - Determine the moved node_id from the item's Qt.UserRole.
    - Call NodeOperations.update_project_id_downtree(node_id) to recompute project_id for the moved node and all descendants.
    - Optionally recompute depth/root_id for that subtree via TreeManager.update_tree_depth_and_root([root_id]) if needed.

- node_operations.py
  - Ensure update_project_id_downtree is robust for arbitrary reparenting (it already walks uptree to nearest Process/Project and propagates downtree). No logic change likely needed, but add logging around drag-and-drop path if a new entry point is added.

- main.py
  - After wiring TreeManager, ensure any new rowsMoved signal handler is connected once and not duplicated.

Verification Steps
- Create two Process/Project roots (Project A and Project B) and a deep subtree under A (parent, child, grandchild).
- Confirm initial project_id for all nodes in the subtree equals A's id.
- Drag-and-drop the parent node from under A to under B in the tree UI.
- Verify in the database (tree table) that project_id for the moved node and all its descendants now equals B's id.
- Run a DOWNTREE(project)-based operation (e.g., Sync or a template using DOWNTREE(project)) and confirm it only sees nodes under the new project and no longer includes nodes from the old project.

Risks/Edge Cases
- If both drag-and-drop and context-menu Move Up/Down are used, ensure project_id updates are not duplicated but remain idempotent.
- Disabling drag-and-drop may affect existing user workflows; if so, prefer the rowsMoved-based fix.

-----------------------------------------------------------------------------
SpecID = B_1_1_G_25F

Four more 'student' Spec Reports are inserted here.

-----------------------------------------------------------------------------------------------

TASK: Grade these Dev Quiz assignments. Score each response 0 to 100 percent (100 = perfect).

Criteria:

   Issue Understanding (25 percent): Accurately restates problem, identifies symptoms vs root cause.
   Root Cause Analysis (30 percent): Logical hypothesis, correctly links to system components.
   Components of Interest (20 percent): Prioritizes relevant files/modules/functions accurately.
   Proposed Changes (15 percent): Specific, actionable modifications... avoids vague suggestions.
   Verification & Risks (10 percent): Clear test steps...  notes edge cases/side effects.

Be strict but fair. Use only provided info - no assumptions.

Output a JSON array, sorted best to worst.

Use the following format for your JSON scores:
--- Artifact Start ---
File: GS_B_1-1.json
```json
[
 {
   "SpecID": "B_1_...",
   "score": 92,
   "brief_justification": "Strong root cause and fixes, minor verification gap"
  }
]
```
--- Artifact End ---

Be sure that you give a grade entry in your response JSON for each of these SpecIDs:

B_1_1_O_51,  B_1_1_G_25F,  B_1_4_O_51,  B_1_2_G_30P, B_1_4_G_25F

The file name for the JSON will be: GS_B_1-1.json.
````

## Appendix E Cost Computation

The Cost column, as found in several tables in this paper, is calculated per published rates (as of Dec 1st 2025) on each vendor's model pricing sheets. The token counts used in the calculation of the cost column are returned from each vendor in their API response JSON ('tokens in' and 'tokens out'). See table below for the pricing that was used in the preparation of this paper.

| Model | Input | Output |
|---|---|---|
| claude-opus-4-5 | 5.00 | 25.00 |
| claude-sonnet-4-5 | 3.00 | 15.00 |
| claude-haiku-4-5 | 1.00 | 5.00 |
| gemini-3-pro-preview | 2.00 | 12.00 |
| gemini-2.5-pro | 1.25 | 10.00 |
| gemini-2.5-flash | 0.30 | 2.50 |
| gpt-5.2 | 1.75 | 14.00 |
| gpt-5.1 | 1.25 | 10.00 |
| gpt-4.1 | 2.00 | 8.00 |
| grok-4-1-fast-reasoning | 0.20 | 0.50 |
| grok-4-1-fast-non-reasoning | 0.20 | 0.50 |
| grok-code-fast-1 | 0.20 | 1.50 |

Table showing the published pricing, in dollars per million, of input and output tokens by model.

Observed *invoiced* amounts across vendors ranged from 2 to 23 times higher than token-based estimates. Costing formulas vary substantially between vendors, making it impossible to determine true cost from the table above alone. Variances may likely be due to minimum charges per API call, separate 'thinking token' charges, caching, tool calls, priority tiers, taxation, etc. - highlighting the economic importance of context compression.

## Appendix F Intellectual Property

**TREEFRAG** and related techniques are protected by seven pending patent applications.

Patent application numbers and titles are available upon reasonable request under standard confidentiality terms.